\documentclass[conference]{IEEEtran}
\IEEEoverridecommandlockouts
\usepackage{cite}
\usepackage{amsmath,amssymb,amsfonts}
\usepackage{algorithmic}
\usepackage{graphicx}
\usepackage{textcomp}
\usepackage{xcolor}
\usepackage[colorlinks,linkcolor=red,citecolor=green]{hyperref}
\usepackage{epstopdf}
\AppendGraphicsExtensions{.tif}
\def\BibTeX{{\rm B\kern-.05em{\sc i\kern-.025em b}\kern-.08em
    T\kern-.1667em\lower.7ex\hbox{E}\kern-.125emX}}
\usepackage{titlesec}

\begin{document}

\title{Self-Supervised Learning for Transparent Object Depth Completion Using Depth from Non-Transparent Objects\\
\thanks{$^{*}$ Corresponding author: Hang Yang. This work was supported by National Natural Science Foundation of China (NSFC) [grant number 62175086]. We thank Associate Professor Luo Shan at King’s College London for his guidance and advice on this research.}
}

\author{
    \IEEEauthorblockN{Xianghui Fan$^{1,2}$, Zhaoyu Chen$^{3}$, Mengyang Pan$^{1,2}$,Anping Deng$^{1,2}$,Hang Yang$^{* 1}$}
    \IEEEauthorblockA{$^1$ Changchun Institute of Optics, Fine Mechanics and Physics, Chinese Academy of Sciences, Changchun, China}
    \IEEEauthorblockA{$^2$ University of Chinese Academy of Sciences, Beijing, China}
    \IEEEauthorblockA{$^3$ Fudan University, Shanghai, China}
    \IEEEauthorblockA{yanghang@ciomp.ac.cn}
}


\maketitle

\begin{abstract}
The perception of transparent objects is one of the well-known challenges in computer vision. Conventional depth sensors have difficulty in sensing the depth of transparent objects due to refraction and reflection of light. Previous research has typically train a neural network to complete the depth acquired by the sensor, and this method can quickly and accurately acquire accurate depth maps of transparent objects. However, previous training relies on a large amount of annotation data for supervision, and the labeling of depth maps is costly. To tackle this challenge, we propose a new self-supervised method for training depth completion networks. Our method simulates the depth deficits of transparent objects within non-transparent regions and utilizes the original depth map as ground truth for supervision. Experiments demonstrate that our method achieves performance comparable to supervised approach, and pre-training with our method can improve the model performance when the training samples are small.
\end{abstract}


\section{Introduction}
Depth information is crucial for a computer's understanding of a scene. Common depth sensors, such as time-of-flight cameras, capture the depth of an object's surface, typically relying on the Lambertian assumption, which assumes that light is diffusely reflected off an object's surface. However, transparent objects, commonly encountered in both everyday life and industrial settings, do not adhere to this assumption, resulting in incomplete depth information when captured by conventional RGB-D cameras\cite{review}.

To address this issue, several methods\cite{CG,A4T,FDCT,segment,TDCNet,TODE,depthgrasp,swindrnet,tim,tcrnet} aim to generate complete depth maps from a single raw RGB-D image. \cite{CG} and \cite{A4T} use global optimization techniques to reconstruct the depth of transparent objects without making external assumptions. However, these methods require the training of additional neural networks to predict surface normals and object edges. In contrast, end-to-end methods have become more popular, as they do not require additional training and offer excellent real-time performance and accuracy.For example, DFNet\cite{transcg} builds a U-Net-like structure using dense blocks and maximizes the utilization of the original depth map by progressively adding it layer by layer. FDCT\cite{FDCT} employs a one-shot aggregate module\cite{FDCTMOULDE} to accelerate the inference process, while fusion branches are designed to leverage the original depth map. Both TODE-Trans\cite{TODE} and SwinDRNet\cite{swindrnet} use Swin Transformer\cite{swin} based backbones, but the former uses a single branch, while the latter incorporates a two-branch fusion design. DualTransNet\cite{segment} introduces an additional sub-network for extracting segmentation features, which are then used to enhance the network’s performance. TDCNet\cite{TDCNet} utilizes a two-branch CNN-Transformer structure to mitigate the loss of information from the original depth map.
   \begin{figure}[!t]
      \setlength{\fboxsep}{0pt}%
      \setlength{\fboxrule}{0pt}%
      \flushright
      \framebox{\includegraphics[width=\linewidth]{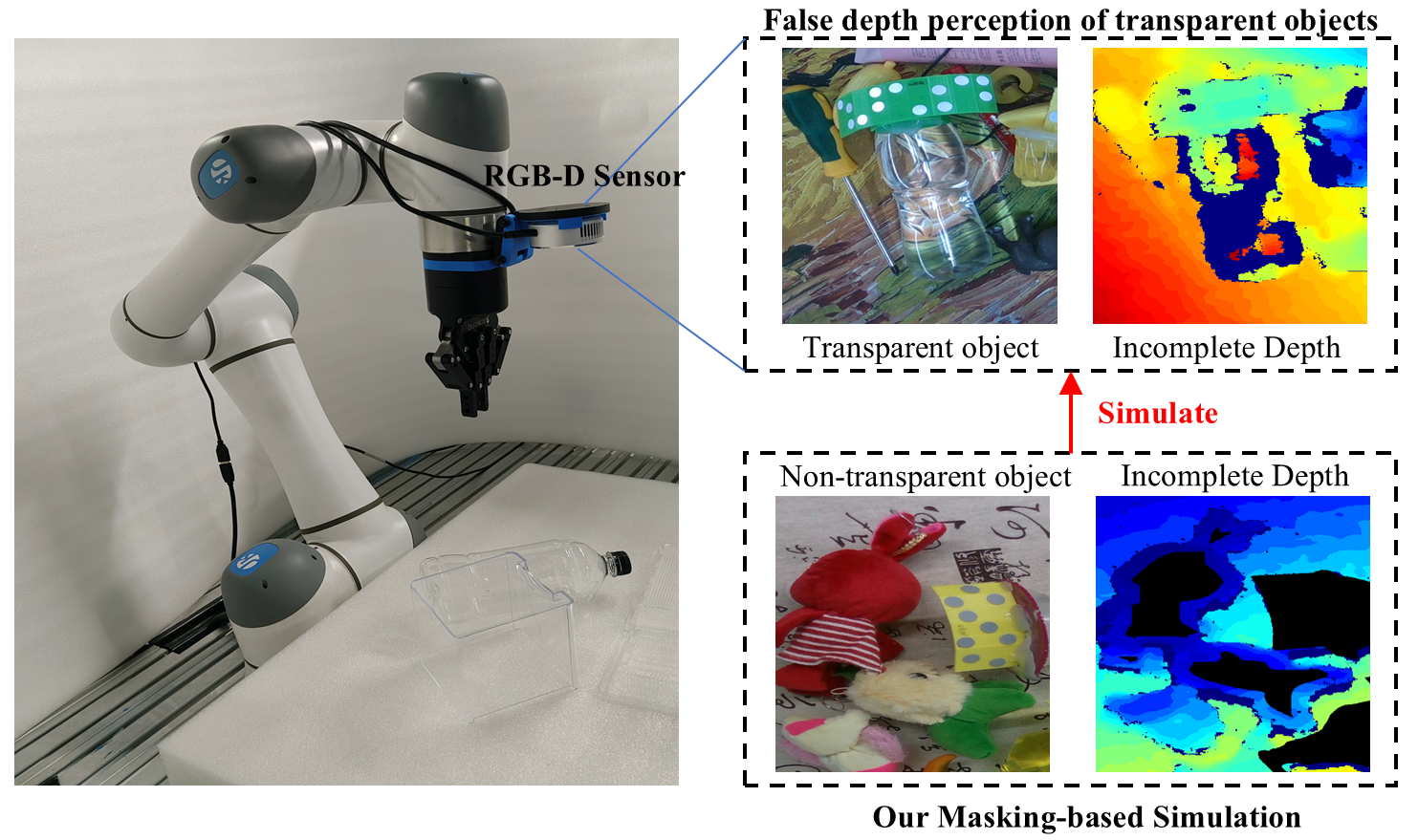}}
      \caption{When the RGB-D sensor captures the depth of a transparent object, a localized depth deficit occurs. We simulate this effect in the non-transparent object region through artificial masking.
      }
      \label{ouridea}
   \end{figure}
However, all the above methods are based on supervised learning, which requires a large amount of labeled data, namely complete depth maps of transparent objects. As mentioned earlier, obtaining such data from conventional depth sensors is challenging, and manually labeling them, as is done in other vision tasks (e.g., object detection, semantic segmentation), is not feasible. These challenges make the use of supervised learning in transparent object depth completion tasks a costly undertaking.

Self-supervised learning\cite{self-survey} seeks to leverage the intrinsic structure and patterns within unlabeled data to automatically generate supervisory data, thereby reducing the dependence on manually labeled data. In indoor depth completion tasks, \cite{self1} and \cite{self2} propose a self-supervised approaches to reduce the reliance on depth ground truth. Both approaches share the idea of artificially simulating further depth deficits on the original depth map, with \cite{self1} employing adaptive sampling and \cite{self2} using the MAE\cite{MAE} stochastic masking strategy. However, both methods\cite{self1,self2} simulate global depth missing across the image, which differs significantly from the localized depth missing of transparent objects, which is confined to the transparent regions.
   \begin{figure*}[tb]
      \setlength{\fboxsep}{0pt}%
      \setlength{\fboxrule}{0pt}%
      \flushright
      \framebox{\includegraphics[width=\linewidth]{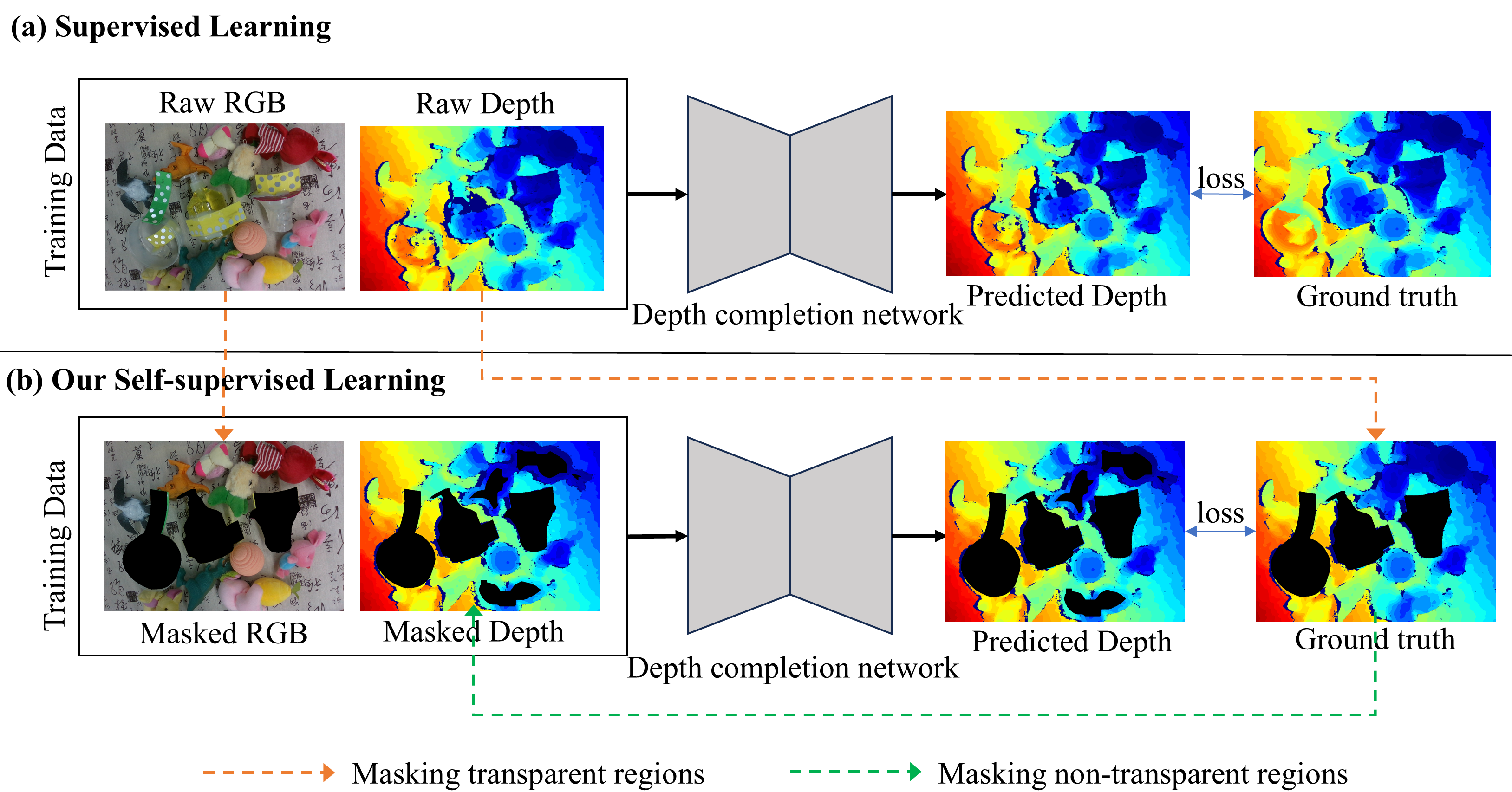}}
      \caption{Pipelines of supervised learning and our self-supervised learning. We use masked input data from the supervised process to perform self-supervised learning, without relying on the full transparent object depth map at any point.
      }
      \label{pipline}
   \end{figure*}
   
 The limitations of the supervised and self-supervised methods mentioned above inspired our approach. As illustrated in Fig. \ref{ouridea}, we aim to model the depth deficits occurring in transparent object regions within non-transparent object regions, effectively creating a task that resembles transparent object depth completion. By training the model on this task, it was able to acquire feature representations that are directly applicable to transparent object depth completion. Specifically, we first use the SAM\cite{SAM} model to generate segmentation masks for transparent and non-transparent targets, and then use these masks to remove the depths of non-transparent targets in the raw depth map. To more closely align with the depth missing pattern of transparent objects, we do not fully mask all the depths of non-transparent objects. The masked depth maps are used as inputs to the model, with the raw depth maps serving as ground truth to supervise the training process. Intuitively, the false depth perception of transparent objects is characterized by local clustering and depth values that deviate from the correct depth in the surrounding areas. Our mask-based simulations align with both of these characteristics. Experiments show that when the model is trained entirely using our self-supervised method, its performance approaches that of supervised learning. Furthermore, pre-training using our method significantly boosts model performance when training data is limited.

Our contributions are as follows:
\begin{itemize}
  \item We propose a self-supervised learning method for transparent object depth completion and introduce a masking strategy that simulates depth deficits in non-transparent object regions, resembling those in transparent object regions.
  \item To the best of our knowledge, our method is the first self-supervised approach specifically designed for the transparent object depth completion task.
  \item Our experiments show that training a model using our self-supervised method achieves performance close to that of supervised learning. Additionally, pre-training with our method significantly enhances model performance.
\end{itemize}

\section{Method}
As shown in Fig. \ref{pipline}, supervised learning uses the original RGB-D image as input and the complete depth map as ground truth, whereas our self-supervised approach derives both the input and ground truth from the original RGB-D image using our masking strategy, and then uses them to train the network. Our masking strategy is designed to simulate a depth completion task for transparent objects, enabling the model to develop specialized feature representation capabilities. Below, we provide a detailed description of our method.
\subsection{Masking Strategy}

\begin{figure}[!t]
      \setlength{\fboxsep}{0pt}%
      \setlength{\fboxrule}{0pt}%
      \flushright
      \framebox{\includegraphics[width=\linewidth]{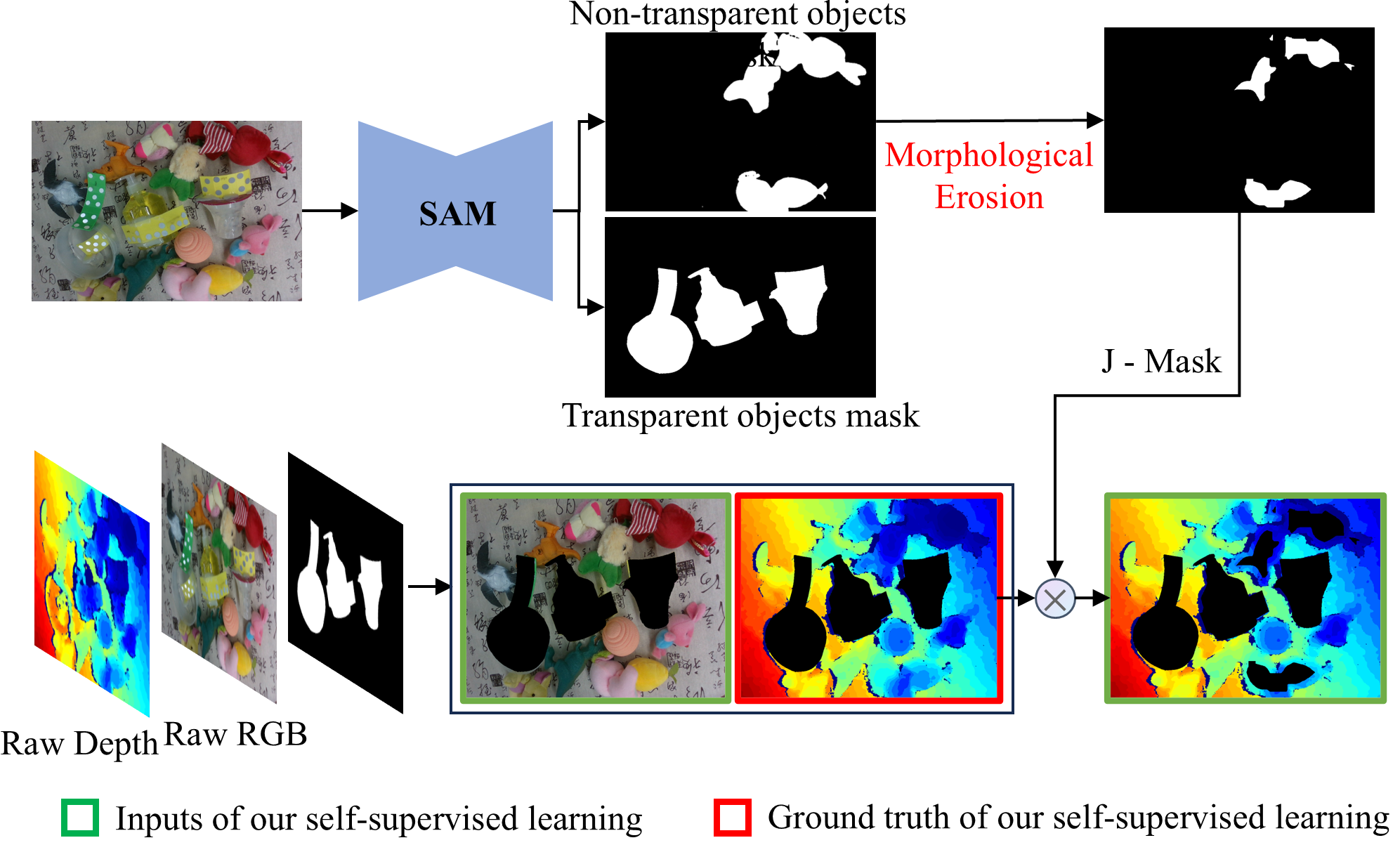}}
      \caption{Our masking strategy
      }
      \label{maskingstrategy}
   \end{figure}

We design a masking strategy to simulate the depth deficit of transparent objects in non-transparent regions. Thus, unlike the MAE method which randomly masks the whole image, our masking strategy only targets the regions of transparent and non-transparent objects in the image. As shown in Fig.~\ref{maskingstrategy}, to localize these object regions, we use the SAM model to obtain semantic segmentation masks for transparent and non-transparent objects. Thanks to the zero-shot capability of the SAM model, the cost of obtaining segmentation masks in this way is low. Subsequently, we use the above masks to cover the depth of non-transparent objects in the depth map. We note that using a mask based on segmentation completely masks the depth of non-transparent objects. Whereas the depth error of transparent objects is not complete, in particular, the depth of the edge contours and the vicinity of the edges can be correctly perceived by the depth sensor. To better model this depth-deficient pattern, we shrink the collected masks of non-transparent objects by one turn, specifically by performing a morphological erosion operation of the mask. Additionally, during self-supervised training, to avoid the influence of transparent objects, we use semantic segmentation masks (also obtained from SAM) to completely obscure transparent objects in both the RGB and depth images. The above process can be described as:
\begin{equation}
M_{non-trans},M_{trans}=SAM(RGB)
\end{equation}
\begin{equation}
M^{\prime}_{non-trans}=M_{non-trans} \textcircled{-} B
\end{equation}
\begin{equation}
M_{final}=J-(M'_{non-trans}\lor M_{trans})
\end{equation}
\begin{equation}
I^{\prime}=(J-M_{trans})\odot RGB
\end{equation}
\begin{equation}
D^{\prime}=M_{final}\odot Depth
\end{equation}
\begin{equation}
D^{\prime}_{gt}=(J-M_{trans})\odot Depth
\end{equation}

Here $SAM$ refers to the SAM model, while $M_{trans}$ and $M_{non-trans}$ represent the masks for transparent and non-transparent objects, respectively. $\textcircled{-}$ denotes the morphological erosion operation, and $B$ is the structuring element used in the erosion, which we set as a $5\times5$ rectangular element. $J$ stands for the all-ones matrix, $\lor$ and $\odot$ represents the logical or computing and Hadamard product. Finally, $I^{\prime}$ and $D^{\prime}$ represent the masked RGB and raw depth maps, which are both used as inputs for self-supervised learning, while $D^{\prime}_{gt}$ is the supervised depth map used for self-supervised learning.

\subsection{Self-supervised Training and Supervised Fine-tuning}
Our self-supervised training process relies solely on the raw depth map and the RGB images processed as described above. During the self-supervised training phase, the goal is to enable the model to recover the whole depth of non-transparent objects using the RGB image (which provides the appearance of the non-transparent object) and the partially damaged depth map. The model we employ is TDCNet\cite{TDCNet}, a two-branch transparent object depth completion network that achieves state-of-the-art performance on datasets like TransCG. In the previous phase, the depth values of the transparent object regions are manually set to 0. Therefore, during the training phase, we avoid computing and accumulating the missing values in the transparent object regions to prevent any potential negative impact on the model's training. Depth loss in the remaining regions are computed, accumulated, and back-propagated normally to reach our goal. Our loss function in self-supervised training is:
\begin{align}
L_{1}=&\frac{1}{N_{x}}\sum_{x}[||D-D_{gt}||^{2}+\alpha(1-\cos<V\times V_{gt}>)], \notag \\
&where \: x\in\{I-Trans\}
\label{eq7}
\end{align}

   \begin{table*}[t]
\centering
\caption{Quantitative results of our full self-supervised learning method versus other methods on the TransCG dataset}
    \resizebox{0.7\linewidth}{!}{
\begin{tabular}{ccccccc}
\hline
\multicolumn{1}{c|}{Methods}    & RMSE↓ & REL↓  & MAE↓  & $\sigma$1.05↑ & $\sigma$1.10↑ & $\sigma$1.25↑ \\ \hline
\multicolumn{7}{c}{Supervised Learning}                                            \\ \hline
\multicolumn{1}{c|}{DFNet\cite{transcg}}      & 0.018 & 0.027 & 0.012 & 83.76  & 95.67  & 99.71  \\
\multicolumn{1}{c|}{CG\cite{CG}}         & 0.054 & 0.083 & 0.037 & 50.48  & 68.68  & 95.28  \\
\multicolumn{1}{c|}{FDCT\cite{FDCT}}       & 0.015 & 0.022 & 0.010  & 88.18  & 97.15  & 99.81  \\
\multicolumn{1}{c|}{TODE-Trans\cite{TODE}} & 0.013 & 0.019 & 0.008 & 90.43  & 97.39  & 99.81  \\
\multicolumn{1}{c|}{TDCNet\cite{TDCNet}}     & 0.012 & 0.017 & 0.008 & 92.37  & 97.98  & 99.81  \\ \hline
\multicolumn{7}{c}{Fully Self-supervised Learning}                                 \\ \hline
\multicolumn{1}{c|}{MAE\cite{MAE}}        & 0.029 & 0.037 & 0.018 & 73.88  & 92.29  & 99.79  \\
\multicolumn{1}{c|}{Ours}       & 0.026 & 0.029 & 0.014 & 81.74  & 94.43  & 99.8   \\ \hline
\end{tabular}
}

\label{full ss table}
\end{table*}
\begin{table*}[!t]
\centering
\caption{Quantitative comparison of fine-tuning results using our self-supervised learning approach versus pre-training with other methods on the TransCG dataset.}
    \resizebox{0.9\linewidth}{!}{
\begin{tabular}{lcccccc}
\hline
\multicolumn{1}{c|}{Methods}                               & RMSE↓          & REL↓           & MAE↓           & $\sigma$1.05↑         & $\sigma$1.10↑         & $\sigma$1.25↑         \\ \hline
\multicolumn{7}{c}{10\% training data   (About 3600 sets of pictures)}                                                                                           \\ \hline
\multicolumn{1}{l|}{Fine-tuning   w/o Pre-training}        & 0.019          & 0.03           & 0.014          & 80.24          & 94.78          & 99.75          \\
\multicolumn{1}{l|}{Fine-tuning w/ Pre-training   by MAE\cite{MAE}}  & 0.019          & 0.031          & 0.014          & 79.69          & 95.14          & \textbf{99.82} \\
\multicolumn{1}{l|}{Fine-tuning w/ Pre-training   by ours} & \textbf{0.018} & \textbf{0.028} & \textbf{0.013} & \textbf{81.79} & \textbf{95.35} & 99.78          \\ \hline
\multicolumn{7}{c}{5\% training data   (About 1800 sets of pictures)}                                                                                            \\ \hline
\multicolumn{1}{l|}{Fine-tuning   w/o Pre-training}        & 0.021          & 0.035          & 0.016          & 75.47          & 93.80           & 99.75          \\
\multicolumn{1}{l|}{Fine-tuning w/ Pre-training   by MAE\cite{MAE}}  & 0.02           & 0.034          & 0.015          & 76.09          & 94.36          & \textbf{99.80}  \\
\multicolumn{1}{l|}{Fine-tuning w/ Pre-training   by ours} & \textbf{0.019} & \textbf{0.03}  & \textbf{0.014} & \textbf{79.21} & \textbf{94.52} & \textbf{99.80}  \\ \hline
\end{tabular}
}
\label{finetune-table}
\end{table*}

In the Eq. \ref{eq7}, $x\in\{I-Trans\}$ represents pixel $x$ in the image excluding the transparent regions, $D$ denotes the predicted depth, and $V$ is the normal vector computed from $D$. $D_{gt}$ and $V_{gt}$ are their corresponding true values, and $<\times>$ is used to compute the cosine similarity. $\alpha$ is a manually set weight, here set by us to $0.1$.

Our supervised fine-tuning process leverages the weights obtained from self-supervised pre-training. Similar to standard supervised training, the fine-tuning process uses the full RGB image and the original depth map as inputs, with the labeled complete depth map serving as the supervision. Loss function in this process is:
\begin{align}
L_{2}=&\frac{1}{N_{x}}\sum_{x}[||D-D_{gt}||^{2}+\alpha(1-\cos<V\times V_{gt}>)], \notag \\
&where \: x\in\{Trans\}
\label{loss2}
\end{align}
\begin{align}
L_{Supervised} = \beta L_{2}+(1-\beta)L_{1}
\label{lossall}
\end{align}

As shown in Eq. \ref{loss2}, we calculate the loss in the transparent region as defined in Eq. \ref{eq7}, and then weight and sum the losses from both the transparent and non-transparent regions. Given that we prioritize depth recovery in the transparent region during the fine-tuning stage, we set the weight $\beta$ to 0.9.

\section{Experiments}
In this section, we present both qualitative and quantitative results of our proposed method on a real-world dataset, covering both fully self-supervised and fine-tuned scenarios using a small amount of data.
\subsection{Experimental Setup}
\textbf{Dataset: }We use the well-known TransCG dataset from Work \cite{transcg}. To the best of our knowledge, TransCG is the largest real-world open-source dataset for depth completion of transparent objects, featuring 57,715 RGB-D images across 130 distinct scenes and 57 different transparent object targets. Notably, the dataset includes scenes with both transparent and non-transparent objects, which is ideal for our self-supervised approach. In contrast, datasets lacking non-transparent objects present challenges for applying our method.
   \begin{figure}[!b]
      \setlength{\fboxsep}{0pt}%
      \setlength{\fboxrule}{0pt}%
      \flushright
      \centering
      \framebox{\includegraphics[width=\linewidth]{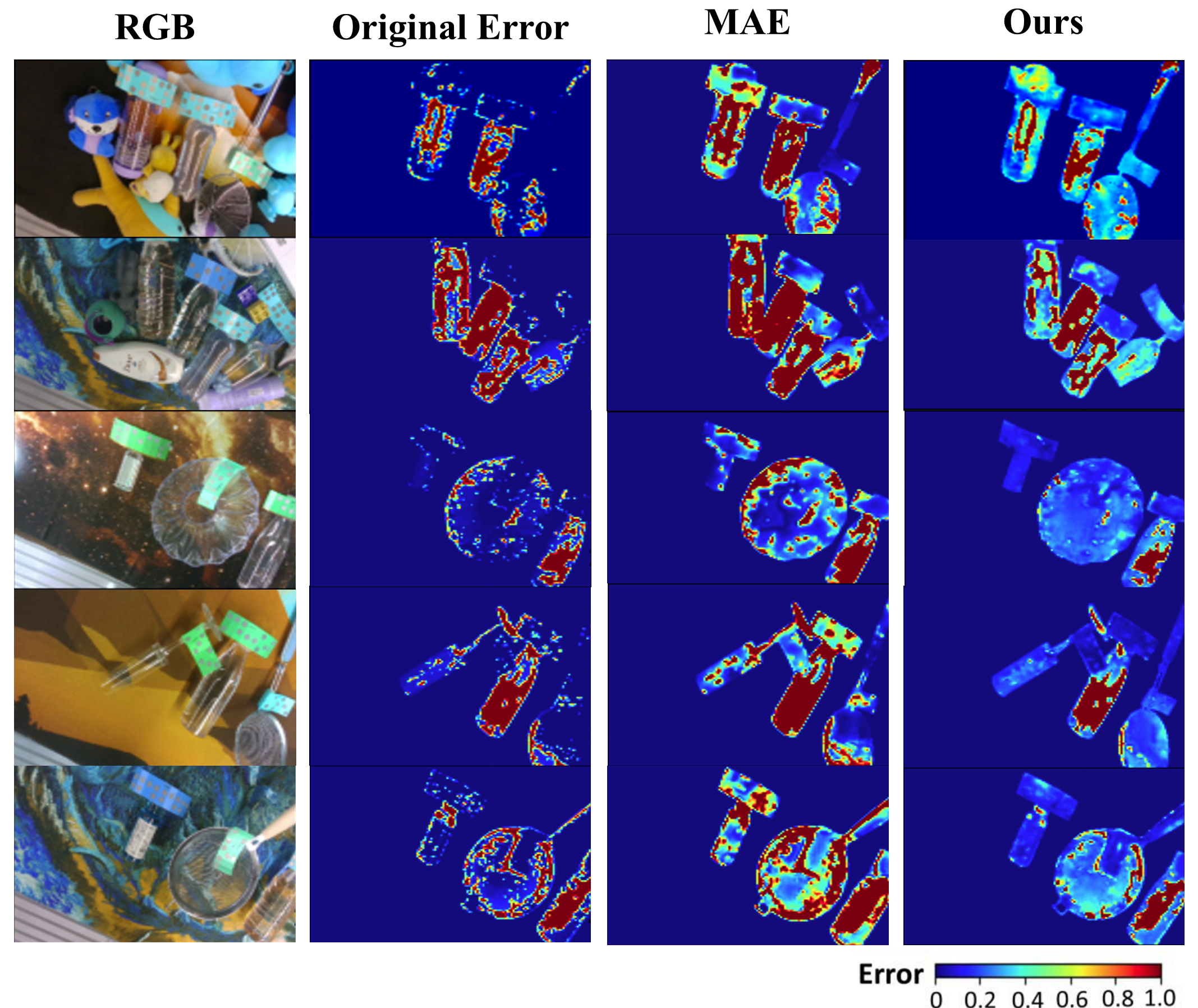}}

      \caption{Qualitative results of our full self-supervised learning method compared to other methods on the TransCG dataset. Each pixel of the error map is calculated by the following relative error: $|d-d^* |/d^*$.The closer the pixel color is to the background, the smaller the relative error, whereas the closer it is to red, the larger the relative error.
      }
      \label{full self}
   \end{figure}

   \begin{figure*}[!t]
      \setlength{\fboxsep}{0pt}%
      \setlength{\fboxrule}{0pt}%
      \flushright
      \framebox{\includegraphics[width=\linewidth]{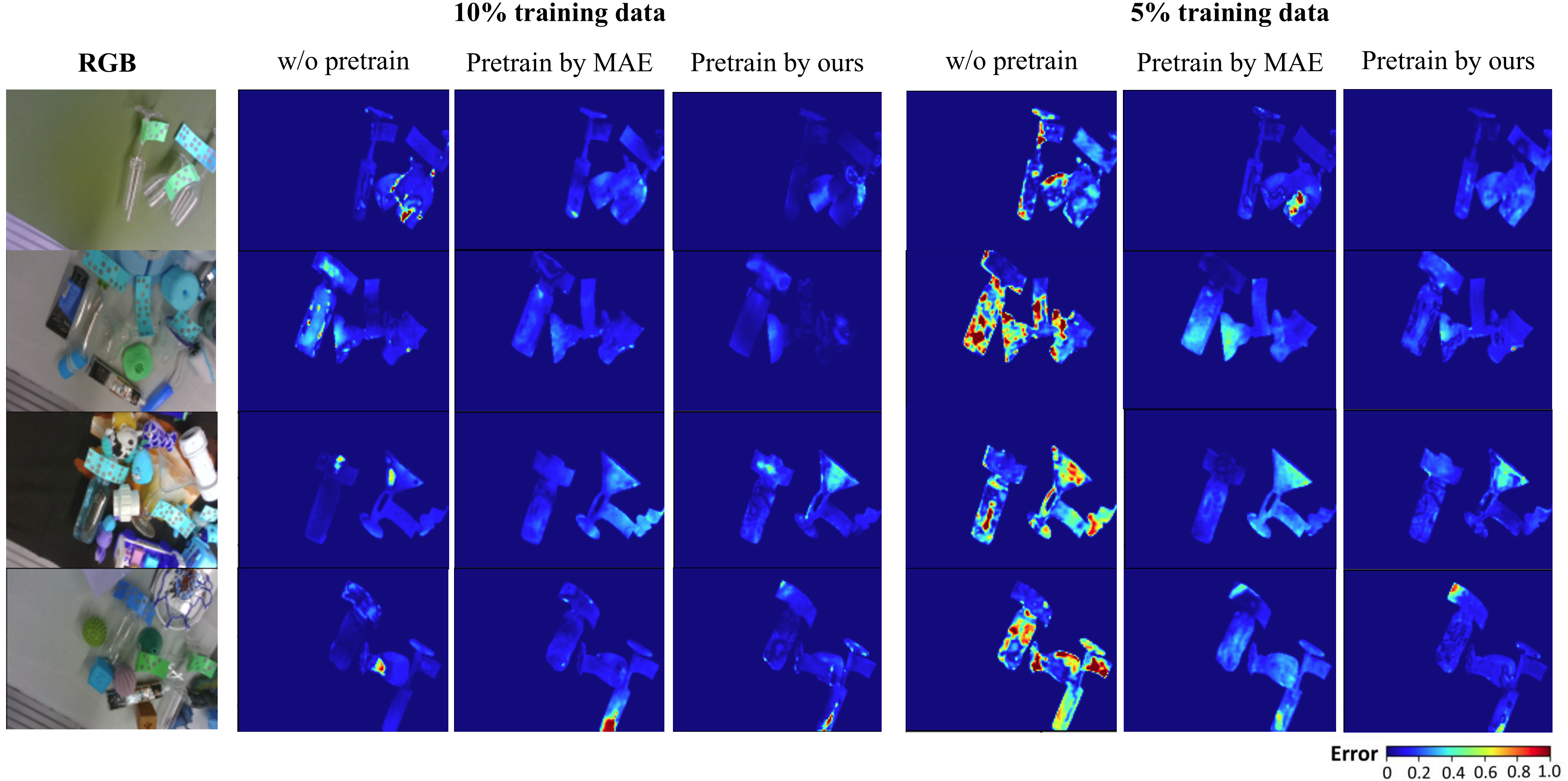}}
      \caption{Qualitative comparison of fine-tuning results using our self-supervised learning approach versus pre-training with other methods on the TransCG dataset. Each pixel of the error map is calculated by the following relative error: $|d-d^* |/d^*$.The closer the pixel color is to the background, the smaller the relative error, whereas the closer it is to red, the larger the relative error.
     }
      \label{finetune}
   \end{figure*}
\textbf{Evaluation Metric: }We evaluate the performance of the depth completion task on the pixels of transparent objects using the following metrics: RMSE (Root Mean Squared Error), REL (Relative Error), MAE (Mean Absolute Error), and Threshold-$\sigma$. Detailed definitions are
\begin{itemize}
\item RMSE: $\sqrt{\frac{1}{|D_{t}|}\sum_{d_{i}\in D_{t}}\|d_{i}-d_{i}^{*}\|^{2}}$
\item REL: $\frac{1}{|D_{t}|}\sum_{d_{i}\in D_{t}}|d_{i}-d_{i}^{*}|/d_{i}^{*}$
\item MAE: $\frac{1}{|D_{t}|}\sum_{d_{i}\in D_{t}}|d_{i}-d_{i}^{*}|$
\item Threshold-$\sigma$: $\max(\frac{d_{i}}{d_{i}^{*}},\frac{d_{i}^{*}}{d_{i}})<\sigma$
\end{itemize}

Where $ D_{t}$ denotes depth values of transparent objects, $d_{i}$ denotes estimated depth value and $d_{i}^{*}$ denotes the corresponding ground truth depth value. The value of $\sigma$ is set to 1.05, 1.10 and 1.25.

\textbf{Implementation Details: }All our experiments were conducted on an Intel I9-14900K CPU and an Nvidia RTX 4090 GPU. During training, we set the batch size to 16 and the input image size to 320 × 240. We used the AdamW optimizer with an initial learning rate of 1e-3, decaying the learning rate by a factor of ten every 15 epochs, for a total of 40 training epochs. Prior to training, we applied several data augmentation techniques, including random flipping, rotation, and the addition of random noise. When we mask the depth map, the depth value of the masked area is set to 0, while the rest of the map remains unchanged. We apply three morphological erosion operations to the non-transparent object mask.

\subsection{Experimental Performance}
\textbf{Fully self-supervised learning: }We train TDCNet on the TransCG dataset using our self-supervised method throughout the entire process, without relying on any ground truth data from the dataset. For a fair comparison with previous methods, our training is conducted exclusively on the training set, although training on the test set would not cause data leakage issues. Table \ref{full ss table} presents our quantitative results. As shown, our self-supervised method achieves 70\% of the performance of previous supervised methods and outperforms certain non-end-to-end methods, such as CG. We also compare our approach with MAE, a self-supervised method that employs randomized masking. It is important to note that the original MAE masking strategy is specific to image tokens, and thus cannot be directly applied to non-Transformer backbones or hierarchical network designs. Adapting it for such cases exceeds the scope of our work. Therefore, we adopt a strategy of randomly masking 75\% of 4×4 pixel blocks in the depth map, which aligns with the original design intent of MAE. To facilitate the comparison of results across different methods, we use error plots to present our qualitative findings. Figure \ref{full self} shows the qualitative results of our approach, where it is clear that our method significantly reduces depth errors, an improvement that MAE cannot achieve.

\textbf{Supervised fine-tuning: }Supervised fine-tuning with a small amount of data following self-supervised pre-training is a more common scenario than full self-supervised training. As shown in Table \ref{finetune-table}, we use the entire TransCG training set for self-supervised pre-training and then perform supervised fine-tuning using 5\% and 10\% of the training set. The quantitative results demonstrate that pre-training with our method offers a significant advantage over both no pre-training and MAE-based pre-training. This advantage becomes more pronounced as the amount of supervised data decreases. We also use error maps to show our qualitative results, the qualitative results shown in Fig. \ref{finetune} further confirm this trend. However, the MAE-based pre-training approach does not result in significant performance gains, which we attribute to its use of a global random mask rather than a localized mask within the target region, as we employed. As a result, the pre-training tasks generated in this manner provide minimal benefit during the fine-tuning phase.

\textbf{Ablation Study}
In our self-supervised approach, we aim to simulate the lack of depth perception for transparent objects by masking the depth of non-transparent objects. As shown in Fig. \ref{maskingstrategy}, our masking is not complete; we preserve the depth around the edges, which we consider more aligned with our objective. This is achieved through morphological erosion operations applied to the semantic segmentation mask. The comparison results in Table \ref{ablations} demonstrate our idea. 
\begin{table}[!t]
\centering
\caption{Ablation experiments with erosion operation}
    \resizebox{\linewidth}{!}{
\begin{tabular}{c|cccccc}
\hline
Methods     & RMSE↓ & REL↓  & MAE↓  & $\sigma$1.05↑ & $\sigma$1.10↑ & $\sigma$1.25↑ \\ \hline
Non-erosion   & 0.031 & 0.048 & 0.026 & 63.62  & 86.90   & 98.92 \\
Erosion     & \textbf{0.026} & \textbf{0.029} & \textbf{0.014} & \textbf{81.74}  & \textbf{94.43}  & \textbf{99.80}  \\ \hline
\end{tabular}
}
\label{ablations}
\vspace{-0.8cm}
\end{table}

\section{Conclusion}
In this paper, we propose a self-supervised method specifically designed for depth completion of transparent objects. Our approach simulates the depth loss pattern of transparent regions by artificially masking the depth in non-transparent regions. The masked depth map is used as input, while the original (unmasked) depth map serves as the ground truth, enabling model training without the need for a complete depth map of the transparent object. Experimental results demonstrate that our method is effective both when applied throughout the training process and when fine-tuned based on our approach.

\textbf{Limitation}
Our self-supervised learning process relies on the depth of non-transparent objects, so the dataset must contain both transparent and non-transparent objects. This requirement makes some datasets incompatible with our method. Additionally, our approach does not perfectly model the missing depth of transparent objects. First, our masking strategy differs from the typical depth missing in transparent objects; we focus on masking depths closer to the center of the target, while the depth loss in transparent objects does not exhibit this pattern. Moreover, the visual disparity between transparent and non-transparent objects still impacts the effectiveness of our self-supervised learning.

\bibliographystyle{IEEEbib}
\bibliography{icme2025references}

\end{document}